# The Dynamic of Belief in the transferable belief model and Specialization-Generalization Matrices.

**Frank Klawonn**
Department of Computer Science
Technical University of Braunschweig
Braunschweig-Germany

**Philippe Smets**
IRIDIA
Université Libre de Bruxelles
Brussels-Belgium

### Abstract:

The fundamental updating process in the transferable belief model is related to the concept of specialization and can be described by a specialization matrix. The degree of belief in the **truth** of a proposition is a degree of *justified support*. The Principle of Minimal Commitment implies that one should never give more support to the truth of a proposition than justified. We show that Dempster's rule of conditioning corresponds essentially to the least committed specialization, and that Dempster's rule of combination results essentially from commutativity requirements. The concept of generalization, dual to the concept of specialization, is described.

## 1. INTRODUCTION.

The dynamic of belief in the Dempster-Shafer theory is achieved by the application of Dempster's rule of conditioning and Dempster's rule of combination. Their appropriateness has been questioned (Levi 1983, Kyburg 1987, Pearl 1990, Voorbracht 1991). But most criticisms result from the fact that the Dempster-Shafer theory is understood as a special form of upper and lower probability theory.

In response to these criticisms, one of us (Smets 1988, 1991b, Smets and Kennes 1990) developed the transferable belief model (TBM), a model to quantify Your[1] belief. The TBM is build without using any concept of probability, hence the criticisms against the use of Dempster's rules do not apply (Smets, 1991a). But there is still the question of explaining why they should be used. This paper presents such a justification.

We present the needed material about the TBM in section 2. In section 3, we introduce the **Principle of Minimal Commitment**. It consists in never giving more support to the elements of the domain of our beliefs than justified. When necessary, it permits us to select the least informative belief function in a set of equally justified belief functions.

In section 4, we study the dynamic components of the TBM, and in particular the expansion process, i.e. the addition of a belief without retracting any old beliefs (Gardenfors 1988). In the TBM, this expansion process is modeled by **specialization** matrices, i.e. operators that reallocate the basic belief masses that quantify our beliefs. The justification of both Dempster's rules is presented in section 5. It proceeds as follows.

When conditioning on an event A, the updated beliefs should give a null plausibility to $\overline{A}$ as $\overline{A}$ becomes impossible. The specialization matrix that satisfies that requirement and that induces the minimal commitment updating leads to an updated belief identical to the one obtained by the application of Dempster's rules of conditioning.

It seems natural to require that combinations of pieces of evidence commute among them and with any conditioning. Dempster's rule of combination is the unique solution satisfying these commutativity requirements.

These results provide justifications for the use of Dempster's rules when beliefs must be expanded by new pieces of evidence.

## 2. THE TRANSFERABLE BELIEF MODEL.

The transferable belief model covers the same domain of application as the subjective probability model except that beliefs are quantified by belief functions, not by probability functions. Let $\Omega$ be a finite set of elements, called the frame of discernment. At most one of the elements of $\Omega$ corresponds to the actual state of affair, but we do not know which one. We can only assess the degrees of belief that this particular element belongs to the various subsets of $\Omega$.

---

[1] 'You' is the agent that entertains the beliefs considered in this presentation.

In the transferable belief model, a unitary amount of belief is postulated, and it is distributed over the subsets of $\Omega$. The part of belief m(A) given to A, called a **basic belief mass** (bbm), quantifies the part of our belief that supports A and does not support any strict subset of A due to lack of information. The bbm m(A) for $A \subseteq \Omega$ are non negative and

$$\sum_{A \subseteq \Omega} m(A) = 1.$$

The function $m: 2^\Omega \to [0,1]$ is called a **basic belief assignment** (bba). The **degree of belief** bel(A) allocated to a subset A of $\Omega$ is the sum of the parts of beliefs m(X) given to subsets X of A that are not subsets of $\overline{A}$ (so $\emptyset$ is not included):

$$\text{bel}(A) = \sum_{\emptyset \neq X \subseteq A} m(X)$$

The **degree of plausibility** pl(A) quantifies the maximum amount of belief that might support A:

$$\text{pl}(A) = \text{bel}(\Omega) - \text{bel}(\overline{A}) = \sum_{X \cap A \neq \emptyset} m(X)$$

The **commonality function** $q: 2^\Omega \to [0,1]$ is defined as

$$q(A) = \sum_{A \subseteq B} m(B)$$

It does not have any practical interpretation but is most useful to represent Dempster's rule of combination (see below).

A **vacuous belief function** is a belief function such that bel(A)=0 $\forall A \neq \Omega$, and bel($\Omega$) = 1. It quantifies our belief in a state of total ignorance. The corresponding bba (where m($\Omega$)=1) and plausibility function (where pl(A) = 1 $\forall A \subseteq \Omega$, $A \neq \emptyset$), are called the **vacuous bba and plausibility function**.

Basic belief assignments, belief functions, plausibility functions and commonality functions are in one to one correspondence (Shafer 1976). The same indexing will be used to show which functions are linked together through that correspondence.

One should note that no normalization is introduced, m($\emptyset$) may be positive, and bel($\Omega$) and pl($\Omega$) may be less than one (Smets 1992). The conclusions still apply when normalization is introduced.

## 3. THE PRINCIPLE OF MINIMAL COMMITMENT.

Let $\Omega$ = {a, b, c}. Suppose You know only that My[1] belief function over $\Omega$ is such that $\text{bel}_{Me}(\{a\})$ = .3 and $\text{bel}_{Me}(\{b,c\})$ = .5, and You do not know the value I give to $\text{bel}_{Me}$ for the other subsets of $\Omega$. Suppose You have no other information on $\Omega$ and You are ready to adopt My belief as Yours. How to build Your belief given these partial constraints. Many belief functions can satisfy them. If you adopt the principle that subsets of $\Omega$ should not get more support than justified, then Your belief on $\Omega$ will be such that $m_{You}(\{a\})$=.3, $m_{You}(\{b,c\})$=.5 and $m_{You}(\{a,b,c\})$=.2. Among all belief functions compatible with the constraints given by known values of $\text{bel}_{Me}$, $\text{bel}_{You}$ is the one that gives the smallest degree of belief to every subsets of $\Omega$. The principle evokes here is called the **Principle of Minimal Commitment**. It is really at the core of the TBM, where degrees of belief are degrees of 'justified' supports.

With un-normalized belief functions, the principle definition is essentially based on the plausibility function. Suppose two plausibility functions $pl_1$ and $pl_2$ such

$$pl_1(A) \leq pl_2(A) \quad \forall A \subseteq \Omega.$$

We say that $pl_2$ is not more committed than $pl_1$ (and less committed if there is at least one strict inequality). The same qualification is extended to their related bba and belief functions. Among all belief functions on $\Omega$, the least committed belief function is the vacuous belief function (i.e. its bbm m($\Omega$)=1).

When expressed with belief functions, the principle becomes:

$$\text{bel}_1(A) + m_1(\emptyset) \geq \text{bel}_2(A) + m_2(\emptyset) \quad \forall A \subseteq \Omega.$$

The concept of 'least commitment' permit the construction of a partial order $\sqsubseteq$ on the set of belief functions (Yager 1986, Dubois and Prade 1986).

We write:

$$pl_1 \sqsubseteq pl_2$$

to denote that $pl_1$ is equal to $pl_2$ or more committed than $pl_2$. By analogy the following notations $m_1 \sqsubseteq m_2$ and $\text{bel}_1 \sqsubseteq \text{bel}_2$ have the same meaning.

The **Principle of Minimal Commitment** consists in selecting the least committed belief function in a set of equally justified belief functions. The principle formalizes the idea that one should never give more support than justified to any subset of $\Omega$. It satisfies a form of skepticism, of uncommitment, of conservatism in the allocation of our belief. In its spirit, it is not far from what the probabilists try to achieve with the maximum entropy principle (see Dubois and Prade 1987, Hsia, 1991, Smets 1991c)

---

[1] 'I' (or 'Me') is an agent different from 'You'.

## 4. THE DYNAMIC OF THE TRANSFERABLE BELIEF MODEL.

In order to build the operators for conditioning and combinations, we formalize the concept of evidential corpus (see Levi 1980 for a similar corpus). We define EC as the **evidential corpus**, i.e. the set of all Your evidence at a given time. Let $EC_1$ and $EC_2$ be two evidential corpus. $EC_1$ and $EC_2$ are said to be equivalent (relative to $\Omega$) if for every piece of evidence E, $EC_1 \cup E$ and $EC_2 \cup E$ induce the same bba on $\Omega$. We say that an evidence E is compatible with EC if $EC \cup E \neq \bot$ where $\bot$ is the contradiction. We say that an evidence E is vacuous (relative to $\Omega$) if EC and $EC \cup E$ are equivalent (relative to $\Omega$) for every EC. A tautology is a typical example of a vacuous evidence.

Among others, change in Your beliefs can be due to expansion, i.e. the addition to EC of an evidence that is compatible with EC. We analyze in detail the expansion process within the TBM where it is represented by two rules: the **Dempster rule of conditioning** and the **Dempster rule of combination**. We explain first the nature of the two rules before introducing the concept of a specialization matrix that will represent the general operator corresponding to the expansion.

Consider first the expansion of Your EC by the fact that all elements of $\overline{A}$ are impossible. It corresponds to the conditioning process on A. As You learn that the elements in $\overline{A}$ are now impossible, the bbm $m(X)$ that was supporting X supports now $X \cap A$, so $m(X)$ is transferred to $X \cap A$. This transfer of bbm results in the unnormalized **Dempster rule of conditioning**. The updated bba $m_A$ and belief function $bel_A$ are:

$$m_A(B) = \sum_{\emptyset \neq Y \subseteq \overline{A}} m(B \cup Y) \quad \text{for } B \subseteq A$$
$$\quad 0 \quad \text{otherwise}$$

and    $bel_A(B) = bel(B \cup \overline{A}) - bel(\overline{A})$  for $B \subseteq \Omega$

Now let EC be Your evidential corpus and let E be a non-vacuous evidence (relative to $\Omega$). We say that E is distinct from EC if there is no non-vacuous evidence F such that $E \models F$ and $EC \models F$. If it were not the case, than F would be a pieces of evidence included in both EC and E and the beliefs they induce on $\Omega$ would result among other from some common underlying evidence F.

Let the evidence E be distinct from the evidential corpus EC. Let $m_0$ and $m_1$ be the bba on $\Omega$ induced by EC and E, respectively. The problem is to build the bba $m_{01}$ induced by $EC \cup E$. Shafer (1976) proposed to compute $m_{01}$ from $m_0$ and $m_1$ by applying **Dempster's rule of combination**. The $\oplus$ symbol is also used to denote this combination. So

$$m_{01} = m_0 \oplus m_1$$

and    $bel_{01} = bel_0 \oplus bel_1$

where $bel_{01}$ is the belief function induced by the bba $m_{01}$ with

$$m_{01}(A) = \sum_{\substack{Y \cap Z = \emptyset \\ Y, Z \subseteq \overline{A}}} m_0(A \cup Y) m_1(A \cup Z) \quad \forall A \subseteq \Omega$$

In particular one has:

$$q_{01}(A) = q_0(A) q_1(A) \quad \forall A \subseteq \Omega$$

Both rules have already received some justifications presented without assuming any underlying probabilistic model. For the conditioning rule, Nguyen and Smets (1991) show that Bel(A|B) is the minimal commitment solution for the belief of the conditional object A|B (Goodman, Nguyen et al., 1991, Goodman, Gupta et al.1991). For the combination rule, Smets (1990), Klawonn and Schwecke (1990), Dubois and Prade (1986), Hajek (1992) provide sets of algebraic requirements that justify it. Another justification based on the concepts of commutative specialization is presented here.

## 5. SPECIALIZATIONS.

The concept of specialization is at the core of the transferable belief model interpretation of the belief allocation by belief functions. Let $m_0$ be a bba induced on $\Omega$ by Your initial EC. The impact of adding a new evidence E to EC induces a change in Your beliefs characterized by a redistribution of the initial $m_0$ such that the bbm $m(A)$ given to the subset A of $\Omega$ is distributed among the subsets of A. Let $s(A,B) \in [0, 1]$ be the proportion of $m(A)$ that flows into $B \subseteq A$ when the new evidence E is added to EC. In order to conserve the whole mass $m(A)$ after this transfer, the $s(A,B)$ must satisfy:

$$\sum_{B \subseteq A} s(A,B) = 1 \quad \forall A \subseteq \Omega$$

As masses can flow only to subsets, then $s(A,B) = 0$ $\forall B \not\subseteq A$. The matrix S of the coefficient $s(A,B)$ for $A, B \subseteq \Omega$ is called a **specialization matrix** on $\Omega$ (see

Yager 1986, Dubois and Prade 1986, Kruse and Schwecke 1990, Moral 1985, Delgado and Moral 1987).

After the expansion of EC by E, the initial bba $m_0$ is transformed into the new bba m such that:

$$m(A) = \sum_{X \subseteq \Omega} s(X,A) \, m_0(X)$$

The bba m is called a specialization of $m_0$. We write $m = m_0 \cdot S$ to represent the specialization of $m_0$ into m by S.

Whenever a bba m is a specialization of a bba $m_0$, then m is at least as committed as $m_0$ (Yager 1986). So $m_0 \cdot S \sqsubseteq m_0$ for any bba $m_0$ and any specialization matrix S (all defined on $\Omega$).

It is easy to show that the effects of both Dempster's rules can be obtained by specialization matrices.

Let $m: 2^\Omega \to [0,1]$ be a bba and $\Sigma$ be the set of specialization matrices on $\Omega$.

1) To obtain **Dempster's rule of conditioning**, let $C \subseteq \Omega$ and let $S_C$ be the specialization matrix such that

$$s_C(A,B) = \begin{cases} 1 & \text{if } B = A \cap C \\ 0 & \text{otherwise} \end{cases}$$

Let $m_C$ be the bba obtained after conditioning the bba m on C by Dempster's rule of conditioning. Then $m_C = m \cdot S_C$. We call $S_C$ the C-conditioning specialization matrix. We define $\Sigma_{cond} \subseteq \Sigma$ as the set of C-conditioning specializations where $C \subseteq \Omega$.

2) To obtain **Dempster's rule of combination**, let m be a bba on $\Omega$ and let $S_m$ be a specialization matrix with coefficients

$$s_m(A,B) = m_A(B) \qquad \forall A,B \subseteq \Omega$$

where $m_A$ is the bba obtained after conditioning the bba m on A by Dempster's rule of conditioning. The coefficients $s_m(A,B)$ satisfy:

$$s_m(\Omega,B) = m(B) \qquad \forall B \subseteq \Omega$$

and $\quad s_m(A,B) = \displaystyle\sum_{X \subseteq \overline{A}} s_m(\Omega, B \cup X) \quad \forall B \subseteq A$

$\qquad\qquad\qquad 0 \qquad\qquad\qquad\text{otherwise}$

Let m' be a bba. One can prove that $m' \oplus m = m' \cdot S_m$. The commutativity of $m' \cdot S_m$ is not obvious but is present as one has also $m' \cdot S_m = m \cdot S_{m'}$.

$S_m$ is called the **Dempsterian specialization matrix associated with m** as it updates any m' on $\Omega$ into $m \oplus m'$. We define $\Sigma_D \subseteq \Sigma$ as the set of the Dempsterian specialization matrices.

In particular, if m is such that $m(C) = 1$ (the bba that corresponds to a conditioning on C), then $S_C = S_m$, so $\Sigma_{cond} \subseteq \Sigma_D$.

We consider that the expansion procedure by the specialization is one of the fundamental ideas for the dynamic part of the transferable belief model. In that model, the bba m(X) given to $X \subseteq \Omega$ corresponds to that part of our belief that supports X without supporting any strict subset of X and which can be transferred to subsets of X if further information justifies it. Therefore we can accept that every expansion is defined by a specialization matrix. We will show that :

1. When conditioning on $A \subseteq \Omega$, $S_A \in \Sigma_{cond}$ is the specialization matrix that induces the least committed beliefs on $\Omega$ derivable by a specialization and such that the updated plausibility given to $\overline{A}$ is 0. The requirement $pl(\overline{A})=0$ after expansion translates the fact that all elements of $\Omega$ in $\overline{A}$ are impossible.

2. $\Sigma_D$ is the largest family of specialization matrices that commute and that includes $\Sigma_{cond}$. The commutativity translates the idea that the expansion by two pieces of evidence should lead to the same result whatever the order with which they are considered. It is classically required for any expansion operator (Gardenfors 1988).

## 6. DEMPSTER'S RULES IN THE VIEW OF SPECIALIZATIONS.

We first show that $S_A \in \Sigma_{cond}$ is the specialization matrix that induces the least committed belief on $\Omega$ derivable by a specialization and such that the updated plausibility given to $\overline{A}$ is 0.

**Theorem 1: Dempster's rule of conditioning.**

Let $C \subseteq \Omega$ and let $\Sigma^* = \{S \in \Sigma \mid \forall m: \text{if } m'=m \cdot S \text{ then } pl'(\overline{C})=0\}$, where pl' denotes the plausibility function induced by $m'=m \cdot S$. Let $S_C$ be the C-conditioning specialization matrix with

$\qquad s_C(A,B)=1 \qquad \text{if } B=A \cap C \text{ and}$
$\qquad s_C(A,B)=0 \qquad \text{otherwise.}$

Then $S_C \in \Sigma^*$ and the bba $m \cdot S_C$ is the least committed bba among the bba $m \cdot S$ for $S \in \Sigma^*$:
$\qquad m \cdot S_C \sqsubseteq m \cdot S \qquad \forall S \in \Sigma^*.$

*Proof.* Let m be a bba over $\Omega$. Let $S \in \Sigma^*$. Let $m' = m \cdot S$. One must have

$$0 = pl'(\overline{C}) = \sum_{B \cap \overline{C} \neq \emptyset} m'(B)$$

$$= \sum_{B \cap \overline{C} \neq \emptyset} \sum_{B \subseteq A} m(A) \cdot s(A,B)$$

It must be true for all bba m on $\Omega$. Therefore $s(A,B)=0$ for all B satisfying $B \not\subseteq C$ and for all $A \subseteq \Omega$. Let S be such a specialization matrix.

$$pl'(D) = \sum_{B \cap D \neq \emptyset} m'(B) = \sum_{B \cap D \neq \emptyset} \sum_{B \subseteq A} m(A) \cdot s(A,B)$$

$$= \sum_{B \subseteq C, B \cap D \neq \emptyset} \sum_{B \subseteq A} m(A) \cdot s(A,B)$$

$$= \sum_{A \cap D \cap C \neq \emptyset} m(A) \sum_{B \subseteq A \cup C, B \cap D \neq \emptyset} s(A,B)$$

$$\leq \sum_{A \cap D \cap C \neq \emptyset} m(A)$$

For any m, let $pl_C$ be the plausibility function induced by $m_C = m \cdot S_C$.

$$pl_C(D) = \sum_{B \cap D \neq \emptyset} m_C(B) = \sum_{B \cap D \neq \emptyset} \sum_{B \subseteq A} m(A) \cdot s_C(A,B)$$

$$= \sum_{A \cap D \cap C \neq \emptyset} m(A)$$

So whatever the bbm m on $\Omega$, $\forall D \subseteq \Omega$, $pl_C(D) \geq pl'(D)$. So $pl_C$ is the least committed plausibility function among those induced by $m \cdot S$ for $S \in \Sigma^*$.   QED

Since Dempster's rule of conditioning is idempotent, so are the conditioning specialization matrices.

**Lemma 1:** The specializations matrices $S \in \Sigma_{cond}$ are idempotent.

We prove that $\Sigma_D$ is the largest set of specialization matrices that commute among themselves and that includes $\Sigma_{cond}$.

**Theorem 2: Dempster's rule of combination.**

Let $S_1 \in \Sigma$ such that for all $C \subseteq \Omega$: $S_1 S_C = S_C S_1$ with $S_C \in \Sigma_{cond}$. Then $S_1 \in \Sigma_D$ holds.

*Proof.* Let $S_{1,C} = S_1 S_C$ with elements $s_{1,C}(A,B)$:

$$s_{1,C}(A,B) = \sum_{D \subseteq \Omega} s_1(A,D) \cdot s_C(D,B)$$

$$= \sum_{X \subseteq \overline{C} \cap A} s_1(A, B \cup X) \cdot s_C(B \cup X, B)$$

$$= \sum_{X \subseteq \overline{C} \cap A} s_1(A, B \cup X) \quad \text{if } B \subseteq C$$
$$= 0 \quad \text{otherwise}$$

Let $S_{C,1} = S_C S_1$ with elements $s_{C,1}(A,B)$:

$$s_{C,1}(A,B) = \sum_{D \subseteq \Omega} s_C(A,D) \cdot s_1(D,B) = s_1(A \cap C, B)$$

Thus for all $A, B, C \subseteq \Omega$, one must have:
$$s_1(A \cap C, B) = \sum_{X \subseteq \overline{C} \cap A} s_1(A, B \cup X) \quad \text{if } B \subseteq C$$
$$= 0 \quad \text{otherwise}$$

If we set $A = \Omega$, we obtain for all $B, C \subseteq \Omega$,

$$s_1(C,B) = \sum_{X \subseteq \overline{C}} s_1(\Omega, B \cup X) \quad \text{if } B \subseteq C$$
$$= 0 \quad \text{otherwise}$$

which implies that $S_1$ is Dempsterian.   QED

Since the Dempster rule of combination is associative and commutative, the Dempsterian specialization matrices commute.

**Theorem 3:** If $S_{m_1}, S_{m_2} \in \Sigma_D$, then $S_{m_1} \cdot S_{m_2} = S_{m_2} \cdot S_{m_1}$.

*Proof:* $m_0 \cdot S_{m_1} \cdot S_{m_2} = (m_0 \oplus m_1) \cdot S_{m_2} = (m_0 \oplus m_1) \oplus m_2 = m_0 \oplus m_1 \oplus m_2 = m_0 \oplus m_2 \oplus m_1 = (m_0 \oplus m_2) \cdot S_{m_1} = m_0 \cdot S_{m_2} \cdot S_{m_1}$   QED

**Note:** Let m be a bba on $\Omega$, with q its related commonality function. Let $S_m$ be the Dempsterian specialization matrix generated by m. Let T be the matrix that transforms a bba into a commonality function, where the elements $t_{A,B}$ of T are:

$$t_{A,B} = 1 \quad \text{if } B \subseteq A \subseteq \Omega$$
$$\quad\quad\quad\ 0 \quad \text{otherwise}$$

It can be shown that 1°) the commonalities $q(A): A \subseteq \Omega$, are the diagonal elements and the eigenvalues of $S_m$ and 2°) the lines of $T^{-1}$ are the eigenvectors of $S_m$[1]. One has the representation: $S_m = T \Lambda T^{-1}$ where $\Lambda$ is a diagonal matrix with elements $\lambda_{A,A} = q(A)$, $A \subseteq \Omega$. Among others, this property may be useful when $m(\Omega) = 0$, in which case the theory of generalized inverses described in matrix calculus can be helpful.

---

[1] The row vector t is an eigenvector of the matrix S if $tS = \lambda t$ where $\lambda$ is a scalar.

## 7. ANOTHER CHARACTERIZATION OF DEMPSTER'S RULE OF COMBINATION.

To understand this theorem, consider that My EC induces the bba $m_0$ on $\Omega$. I want to build a specialization matrix S ready to combine the impact of any new evidence distinct from EC, i.e. to specialize any new bba m on $\Omega$. Let $\Sigma_0$ be the set of the specialization matrices S such that the application of S on any m on $\Omega$ would be at least as committed as $m_0$. One of the elements of $\Sigma_0$ is $S_{m_0}$, the Dempsterian specialization matrix generated by $m_0$. For any bba m on $\Omega$, the bba $m \cdot S_{m_0}$ is less committed than the other bba $m \cdot S$ that could be obtained by the use of the other specialization matrices S of $\Sigma_0$. Dempster's rule of combination satisfies that form of minimal commitment.

**Theorem 4**: Let $m_0$ be a bba on $\Omega$. Let $\Sigma_0 = \{S \in \Sigma \mid \forall m: m \cdot S \sqsubseteq m_0\}$. Then $S_{m_0} \in \Sigma_0$ and $\forall S \in \Sigma_0, \forall m, m \cdot S \sqsubseteq m \cdot S_{m_0}$.

*Proof.* $S_{m_0} \in \Sigma_0$ holds, since $m \cdot S_{m_0} = m \oplus m_0 \sqsubseteq m_0$. Let $S_0 = S_{m_0}$. Let $S \in \Sigma_0$. For $A \subseteq \Omega$, we define the bba $m_S^{(A)}$ by $m_S^{(A)}(B) = s(A,B)$ with corresponding plausibility function $pl_S^{(A)}$.

Let m be the bba with $m(A)=1$.

$m \cdot S = m_S^{(A)}$ and therefore $m_S^{(A)} \sqsubseteq m_0$ implying

$m_S^{(A)} = m_S^{(A)}(.|A) \sqsubseteq m_0(.|A) = m_{S_0}^{(A)}$. (*)

Now let m be an arbitrary bba on $\Omega$. Let pl' be the plausibility function corresponding to $m \cdot S$.

$$pl'(A) = \sum_{B \cap A \neq \emptyset} \sum_{B \subseteq C} m(C) \cdot s(C,B)$$

$$= \sum_{C \cap A \neq \emptyset} \sum_{B \subseteq C, B \cap A \neq \emptyset} m(C) \cdot s(C,B)$$

$$= \sum_{C \cap A \neq \emptyset} m(C) \cdot pl_S^{(C)}(A)$$

$$\leq \sum_{C \cap A \neq \emptyset} m(C) \cdot pl_{S_0}^{(C)}(A) = pl''(A)$$

where pl" denotes the plausibility function corresponding to $m_0 \cdot S_0$ and the inequality results from (*).    QED

## 8. GENERALIZATION TO CONTRACTION OF BELIEFS AND DISJUNCTIVE COMBINATIONS.

The presentation centered around the expansion process. But it can be adapted in a direct way to the inverse process of contraction, i.e. the change in belief where a belief is eliminated from the evidential corpus EC. The specialization matrices encountered in the expansion processes are replaced by de-specialization matrices. The de-specialization matrices are the (matricial) inverse of the specialization matrices. In contrast with the specialization matrices, they should not be applied to any bba, but only to a subset of bba that depends on the de-specialization matrix. This constraint reflects the fact that one can not eliminate from EC an evidence that had not been previously added to EC.

Pieces of evidence are combined conjunctively in the expansion: the aim is to compute the belief function $bel_E$ and F induced by the conjunction 'E and F' of two pieces of evidence E and F from the belief functions $bel_E$ and $bel_F$ induced by each piece of evidence. In the retraction, the process can be seen as a special form of conjunctive combination. Eliminating evidence E from the EC that contains only the two distinct pieces of evidence E and F consists in adding the information 'eliminate E'. It aims at computing $bel_F$ from $bel_{E \text{ and } F}$ and $bel_E$ (in practice an easy task when commonality functions are considered as $\forall A \subseteq \Omega \, q_F(A) = q_{E \text{ and } F}(A) / q_E(A)$).

One can define the processes of disjunctive combination of evidence that is dual to the conjunctive combination. The typical example is illustrated by the case where Your initial $EC_0$ is vacuous, where You know $bel_E$ and $bel_F$, and you can add only 'E or F' to $EC_0$. You must established $bel_{E \text{ or } F}$ from $bel_E$ and $bel_F$. The disjunctive rule of combination has been defined and justified in Smets (1991c), the major result being that the product of bbm $m_E(A)$ and $m_F(B)$ is transferred to $A \cup B$ in $m_{E \text{ or } F}$, and $\forall A \subseteq \Omega \, bel_{E \text{ or } F}(A) = bel_E(A) \, bel_F(A)$.

The dual in the disjunctive context of the specialization matrix encountered in the conjunctive context is called a generalization matrix G. It is also a stochastic matrix where the elements $g(A,B)$ are null $\forall A \not\subseteq B \subseteq \Omega$. It represents the fact that a bbm $m(A)$ is distributed among the supersets of A. The dual of the conditioning process is called the enlargement process. Let m' be the enlargement on A of a bba m. Then the conditioning of m' on any subset $X \cup Y$ of $\Omega$ with $X \subseteq \overline{A}$ and $Y \subseteq A$ will give the same result whatever $Y \subseteq A$. The enlargement transforms m into m' where the elements of A become 'indiscernible' (in that no subset of A receives a support that is not given to the whole A).

The next table summarizes the dual relations.

| evidence combination | conjunctive | disjunctive |
|---|---|---|
| from $bel_E$ and $bel_F$ | $bel_E$ and F | $bel_E$ or F |
| expansion matrices | specialization | generalization |
| retraction matrices | de-specialization | de-generalization |
| single fact | conditioning | enlargement |
|  | de-conditioning | de-enlargement |
| combination | conjunctive rule | disjunctive rule |

Finally it can also be shown that the dual and the inverse relations are in one-to-one correspondence. To any de-specialization matrix corresponds a generalization matrix, and vice versa. So we do not have four processes, but only two. Which of each pair of equivalent approaches is more convenient depends on the context.

## 9. CONCLUSIONS.

We provide justifications for Dempster's rule of conditioning and Dempster's rule of combination based on the following assumptions:

1. the expansion of any evidential corpus EC by an evidence E induces an updating from a prior bba $m_0$ into a posterior bba $m_1$ such that $m_1$ is a specialization of $m_0$.

2. if the evidence E says that all elements in $\overline{A} \subseteq \Omega$ are impossible, then the specialization $m_1$ should be such that $pl_1(\overline{A})=0$, and the least committed specialization $m_1$ is the one obtained by applying Dempster's rule of conditioning on $m_0$.

3. the result of the expansion of Your EC by two pieces of evidence should be independent of the order with which the two pieces of evidence are taken in consideration. Therefore commutativity is required for the specialization matrices that represent these expansions, including those describing the conditioning process. The only specialization matrices satisfying these commutativity requirements are such that the result of their application is the same as the one obtained by the application of Dempster's rule of combination.

### Acknowledgment.

This research has been supported by the ESPRIT II Basic Research Project 3085 (DRUMS) which is funded by the European Communities.